\documentclass{article}

\usepackage{arxiv}

\usepackage[utf8]{inputenc} 
\usepackage[T1]{fontenc}    
\usepackage{hyperref}       
\usepackage{url}            
\usepackage{booktabs}       
\usepackage{amsfonts}       
\usepackage{nicefrac}       
\usepackage{microtype}      
\usepackage{lipsum}
\usepackage{graphicx}
\usepackage{amsmath}
\usepackage{algorithm}
\usepackage{algpseudocode}
\usepackage{rotating}
\graphicspath{ {./images/} }  

\title{A Fuzzy Time Series Based Model Using Particle Swarm Optimization and Weighted Rules }
\author{
 Daniel Ortiz-Arroyo  \\
  Department of Energy \thanks{Author acknowledges the contribution of Jens Runi Poulsen in this research}\\
  \texttt{doa@et.aau.dk} }

\begin{document}
\maketitle

\begin{abstract}
During the last decades, a myriad of fuzzy time series models have been proposed in scientific literature. Among the most accurate models found in fuzzy time series, the high-order ones are the most accurate. The research described in this paper tackles three potential limitations associated with the application of high-order fuzzy time series models. To begin with, the adequacy of forecast rules lacks consistency. Secondly, as the model's order increases, data utilization diminishes. Thirdly, the uniformity of forecast rules proves to be highly contingent on the chosen interval partitions. To address these likely drawbacks, we introduce a novel model based on fuzzy time series that amalgamates the principles of particle swarm optimization (PSO) and weighted summation. Our results show that our approach models accurately the time series in comparison with previous methods.

\end{abstract}


\section{Introduction}
Fuzzy time series modeling is a technique that models time series using fuzzy logic \cite{KlirJuan1,Zadeh1}. During the last two decades a large number of fuzzy time series models have been proposed to model a variety of forecast problems, such as university enrollments \cite{TsaiYu1,SongChissom1,SongChissom2,Chen1,Chen2,SullivanWoodall1,Sing1,Sing2,LiCheng1,ChenHsu1,ChenChung1,LeeChou1,Tsaur1,ChenHwangLee1,LiChen1,Cheng1,Huarng1,Cheng2,Kuo1,Stevenson1}, stock market indexes \cite{ChenChengTeoh1,HuarngYuHsu1,JilaniBurney1,HuarngYu2,Huarng2,HuarngYu3,Yu1,Cheng2}, temperature prediction \cite{LiChengLin1,ChenHwang1,Huarng1}, car road accidents \cite{JilianiBurneyArdil1}, IT-project costs \cite{Cheng1} and annual rice production \cite{Sing1,Sing2}.

Originally, the concept of fuzzy time series was proposed by Song and Chissom \cite{SongChissom1,SongChissom2,SongChissom3} in a paper series to serve as a framework for forecast modeling. They proposed the time-variant and the time-invariant model to deal with the problem of forecasting student enrollments at the University of Alabama. Subsequently, Chen \cite{Chen1} introduced a simplified procedure that reduced the high computational overhead of its predecessors. In a subsequent investigation, Chen \cite{Chen2} expanded upon his prior research outlined in \cite{Chen1}, wherein he introduced a high-order fuzzy time series model. This extended model exhibited notably superior performance compared to its first-order counterpart. More recently, Chen and Chung \cite{ChenChung1} further built upon Chen's work detailed in \cite{Chen1} and \cite{Chen2}. In their latest work, they present a forecasting method that leverages high-order fuzzy time series and employs a genetic algorithm paradigm to collectively adjust interval lengths, aiming to enhance forecast accuracy. A somewhat analogous approach was proposed by Kuo et al \cite{Kuo1}, where particle swarm optimization (PSO) is used in a similar manner. Up to this point, these two models have yielded the most promising results when applied to the enrollment dataset at the University of Alabama. Nevertheless, there remain three potential shortcomings associated with high-order models that require attention. Firstly, the forecast accuracy remains unsatisfactory. Secondly, as the order increases, data utilization diminishes. Thirdly, forecast accuracy proves to be highly sensitive to the chosen interval partitions. To address these issues, we introduce a novel fuzzy time series-based model that combines particle swarm optimization (PSO) with weighted fuzzy rules. The fuzzification method was originally proposed in \cite{ortiz2018weighted}, in this paper, we provide a more detailed description of the model with examples. It must be noted that in this paper we do not consider the forecasting problem for time series but the creation of an accurate fuzzy time series model capable of approximating the time series with precision.

The remaining part of the paper is organized as follows. Section \ref{sec:pso} reviews the concept of PSO. Section \ref{sec:algorithm} provides an overall description of our proposed algorithm. Section \ref{sec:forecastingenrollments} demonstrates how the proposed algorithm is used to model a short time series. Section \ref{sec:results} presents experimental results. Finally, in section \ref{sec:conclusions}, we provide our conclusion.

\section{Particle Swarm Optimization (PSO)}
\label{sec:pso}
Particle Swarm Optimization (PSO) \cite{KennedyEberhart1,KennedyEberhart2} is an optimization method that draws inspiration from the collective behaviors observed in bird flocks and fish schools. In PSO, bird flocks are simulated as particle swarms navigating through a virtual search space in search of the optimal solution. Each particle is assigned a fitness value, which is assessed against a fitness function that needs to be optimized. The motion of each particle is influenced by a velocity parameter. In each iteration, the particles move randomly within a confined area, but their individual movements are guided toward the particle that is closest to the optimal solution. Each particle retains knowledge of its own best position (the best solution it has discovered) as well as the global best position (the best solution found by any particle within the group). The parameters are continually updated each time a new best position is identified, leading to an evolving solution as each particle adjusts its position.

PSO starts with the initialization of a collection of randomly generated particles, each representing a candidate solution. Subsequently, an iterative process is used to enhance the existing set of solutions. In the course of each iteration, every particle generates new solutions, which are individually assessed against (1) the particle's own best solution achieved in prior iterations and (2) the best solution currently identified by any particle within the entire swarm. Each candidate solution is represented as a position. When a particle discovers a position superior to its current best-known position, its personal best position is then updated. Moreover, if the new personal best position surpasses the current global best position, the global best position is also adjusted. Once the evaluation phase concludes, each particle updates both its velocity and position using the following equations:

\begin{equation}
\label{eq9}
	v_{i} = \omega v_{i} + c_{1} r_{1}(\hat{x}_{i} - x_{j}) + c_{1}r_{2}(\hat{g} - x_{j})
\end{equation}
\begin{center}
and
\end{center}
\begin{equation}
\label{eq10}
x_{j} = x_{j} + v_{i}
\end{equation}

\noindent where 

\begin{itemize}
	\item $v_{i}$ is the velocity of particle $p_{i}$ and is limited to $[-V_{max}, V_{max}]$ where $V_{max}$ is a user-defined constant.
\item $\omega$ is an inertial weight coefficient.
\item $\hat{x}_{i}$ is the current personal best position.
\item $x_{j}$ is the current position.
\item $\hat{g}$ is the global best position.
\item $c_{1}$ and $c_{2}$ are user-defined constants saying how much the particle is directed towards good positions. They affect how much the particle's local best and global best influence its movement. Generally $c_{1}$ and $c_{2}$ are set to 2. 
\item $r_{1}$ and $r_{2}$ are randomly generated numbers between 0 and 1.
\end{itemize}

Note that the velocity controls the motion of each particle. The speed of convergence can be adjusted by the inertial weight coefficient and the constants $c_{1}$, $c_{2}$. Whenever computed velocity exceeds its user-defined boundaries, the computed results will be replaced by either $-V_{min}$ or $V_{max}$. 


\begin{algorithm}
\caption{\footnotesize The running procedure of the PSO Algorithm}
\begin{algorithmic}
\footnotesize
\For{all particles}
\State initialize velocities and positions
\EndFor
\While{stopping criteria is unsatisfied}
\For{each particle}
\State compute velocities by equation \ref{eq9}
\State increment positions by equation \ref{eq10}
\If{current fitness value is better than current local best value}
\State update local best positions
\EndIf
\If{current fitness value is better than current global best value}
\State update global best positions
\EndIf
\EndFor
\EndWhile
\label{algorithm2}
\end{algorithmic}
\end{algorithm}

\section{Algorithm Overview}
\label{sec:algorithm}
\begin{figure}[ht]
	\centering
		\includegraphics[scale=0.8]{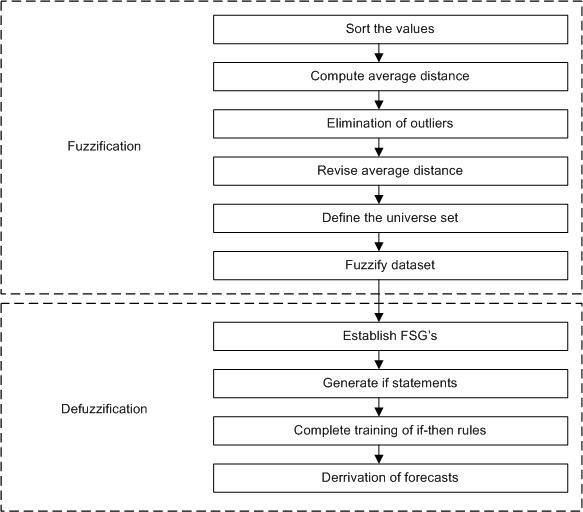}
	\caption{Overall algorithm structure}
	\label{fig4}
\end{figure}

In this study, we have developed a training algorithm, which will be discussed in detail in the subsequent sections. The training phase will utilize the short enrollment dataset from the University of Alabama \cite{SongChissom1} as input. The output of this phase consists of forecast enrollments, which will be employed for evaluating the performance in comparison to related studies, as outlined in Section \ref{sec:results}. The comprehensive structure of the algorithm is depicted in Figure \ref{fig4}.

The algorithm can divided into two main components, fuzzification and defuzzification. These are highlighted by the dashed areas. Both of these components are decoupled which implies that they can function independently of each other and thus can be used in combination with other alternatives.
The fuzzification component is further decomposed into a six-step process where the first four steps are data preprocessing functions. The fuzzification task itself comprises the last two steps. Ultimately the goal of this phase is to generate a series of fuzzy sets or interval partitions and to establish associations between the fuzzy sets and the dataset values. After the fuzzification process is completed, data is further processed by the defuzzification component. During this phase, the fuzzy sets generated in the previous phase, are grouped into patterns and transformed into forecast rules. Finally, forecasts are computed by matching the if-then rules with equivalent patterns in the enrollment dataset.

\section{Forecasting enrollments with the proposed method}
\label{sec:forecastingenrollments}

\subsection{Fuzzifying the Data}
\label{sec:fuzzyfication}
The fuzzification algorithm described here is a further modification of the trapezoid fuzzification approach proposed by Cheng et al in \cite{Cheng1}. It differs in the way that it doesn't require the number of sets to be submitted in advance. Instead, the algorithm determines the number of sets based on the variations in data. The advantage of this approach is that the fuzzification process can be carried out automatically. The proposed procedure can be described as a six-step process:

\begin{enumerate}
	\item Sort the values in ascending order.
	\item Compute the average distance between any two consecutive values and the standard deviation.
	\item Eliminate outliers.
	\item Compute the revised average distance between any two remaining consecutive values.
	\item Define the universe of discourse.
	\item Fuzzify the dataset.
\end{enumerate}

First, the values in the historical dataset are sorted in ascending order. Then the average distance between any two consecutive values in the sorted dataset is computed and the corresponding standard deviation. The average distance is given by the equation:         
  
\begin{equation}                  	
	AD(x_{i} ,\ldots, x_{n}) = \frac{1}{n-1} \sum_{i=1}^{n-1} |x_{p(i)} - x_{p(i+1)}|,
\label{eq1}
\end{equation}
where $p$ is a permutation that orders the values ascendantly: $x_{p(i)} \leq x_{p(i+1)}$. The standard deviation is computed as

\begin{equation}
\sigma_{AD} = \sqrt {\frac{1}{n} \sum_{i = 1}^{n} (x_{i} - AD)^2}.
\label{eq2}
\end{equation}	

Both the average distance and standard deviation are used in step 3 to define outliers in the sorted dataset. Outliers are values that are either abnormally high or abnormally low. These are eliminated from the sorted dataset in order to reduce the impact of distorting elements on the average distance value. An outlier, in this context, is defined as a value less than or larger than one standard deviation from the average. After the elimination process is completed, a revised average distance value is computed for the remaining values in the sorted dataset, as in step 2. The revised average distance, obtained in step 4, is used in steps 5 and 6 to partition the universe of discourse into a series of trapezoidal fuzzy sets. In step 5, the universe of discourse is determined. Its lower and upper bound is determined by locating the largest and lowest values in the dataset and augmenting these by (1) subtracting the revised average distance from the lowest value and (2) adding the revised average distance to the highest value. More formally, if $D_{max}$ and $D_{min}$ are the highest and lowest values in the dataset, respectively, and $AD_{R}$ is the revised average distance, the universe of discourse, $U$, can be defined as $U = [D_{min} - {AD}_{R},D_{max} + AD_{R}]$.

When $U$ has been determined, fuzzy subsets can be defined on $U$. Since subsets are represented by trapezoidal functions, the membership degree, for a given function, $\mu_{A}$, and a given value, $x$, is obtained by the equation:

\begin{equation}
\mu _{A} =
\begin{cases}
{\dfrac{ x-a_{1}}{a_{2}-a_{1}}}, & a_{1} \leq x \leq a_{2}\\
1, & a_{2} \leq x \leq a_{3}\\
{\dfrac{a_{4}-x}{a_{4}-a_{3}}}, & a_{3} \leq x \leq a_{4}\\
0, &\text{otherwise.}
\end{cases}
\label{eq3}
\end{equation}

Prior to the fuzzification of data, we need to know the number of subsets to be defined on $U$. The number of sets, $n$, is computed by

\begin{equation}
n = \frac{R - S}{2S} ,
\label{eq4}
\end{equation}

where $R$ denotes the range of the universe set and $S$ denotes the segment length. Equation \ref{eq4} originates from the fact that we know the following about $S$:

\begin{equation}
S= \frac{R}{2n+1}.
\label{eq5}
\end{equation}

\begin{figure}[ht]
	\centering
		\includegraphics{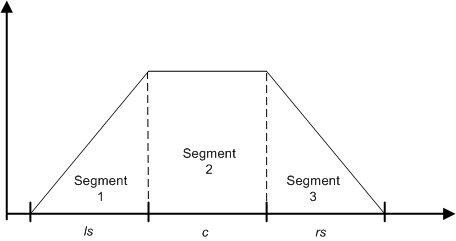}
	\caption{The segments of a trapezoidal fuzzy number}
	\label{fig1}
\end{figure}

The range, $R$, is computed by

\begin{equation}
R = UB - LB,
\label{eq6}
\end{equation}

where $UB$ and $LB$ respectively denote the upper bound and lower bound of $U$. The segment length, $S$, equals the average revised distance, $AD_{R}$, which in turn constitutes the length of left spread $(ls)$, core $(c)$, and right spread $(rs)$ of the membership function (see fig. \ref{fig1}). That is $ls = AD_{R}, c = AD_{R}$ and $rs = AD_{R}$ .

In short, the task here is to decide how many fuzzy sets to generate when the length of each segment, $S$, and the range, $R$, are known. When the number of sets has been computed, the sets can be defined on $U$ and data can be fuzzified which completes the final step of the algorithm.

\subsubsection{An Example}
\label{subsec:anexample}

In the following example, we will fuzzify the first four years (1971 - 1974) of student enrollment at the University of Alabama. The values to be fuzzified are 13055, 13563, 13867, and 14696. Since the sequence of values already appears in ascending order, the sorting part is omitted. The average distance and the standard deviation are respectively computed as 

\[AD = \frac{|13055-13563|+|13563-13867|+|13867-14696|}{3} = 547\]
\begin{center} 
and 
\end{center}
\[
\sigma_{AD} = \sqrt{\frac{(508-547)^2 + (304-547)^2 + (829-547)^2}{3}} = 216.1 \approx 216.
\]

Next, possible outliers are eliminated. Recall that outliers include the values less than or larger than one standard deviation from $AD$. This means only the values satisfying the condition: 

\[547 - 216 \leq x \leq 547 + 216,\] 
are taken into consideration when computing the revised average distance. In this case, only one of the three values satisfies the above condition, namely 508. Thus the revised average distance, $AD_{R}$, and the segment length, $S$, equals 508. At this point, steps 1 - 4 are completed. Prior to defining the universe set, $U$, we need to determine the lower bound and the upper bound $(UB)$ of $U$. Following equation \ref{eq6}, $LB$ and $UP$ are computed as

\begin{align}
LB = 13055 - 508 & = 12547 \notag \\
UB = 14696 + 508 & = 15204.	\notag
\end{align}

Hence $U = [12547, 15204]$. The range, $R$, is computed as the difference between $UB$ and $LB$. Hence we get $15204 - 12547 = 2657$. Finally, the number of sets, $n$, is computed as 

\[
n = \frac{2657 - 508}{2 \cdot 508} = 2.12 \approx 2.
\]

\begin{figure}[ht]
	\centering
		\includegraphics[scale=0.7]{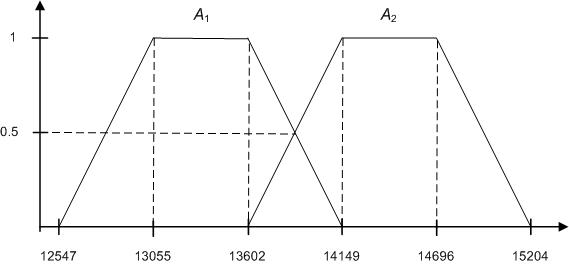}
	\caption{Generated membership functions.}
	\label{fig2}
\end{figure}

\noindent Knowing the universe of discourse and the parameters of $n$, $R$ and $S$, the fuzzy sets are generated as shown in figure \ref{fig2} and table \ref{tab1}.

\begin{table}[ht]
	\centering
	\footnotesize
		\begin{tabular}{ccc}
		\hline 
		\textbf{Fuzzy set} & \textbf{Trapezoidal fuzzy number (a,b,c,d)} & \textbf{Crisp interval} \\
		\hline 
		$A_{1}$ & (12547,13055,13602,14149) & $\mu_{1} = [13055,13602]$ \\
		$A_{2}$ & (13602,14149,14696,15402) & $\mu_{1} = [14149,14696]$ \\
		\hline			
		\end{tabular}
		\caption{Fuzzifying the first four years of enrollment.}				
	\label{tab1}
\end{table}

Note the difference between the points $a$, $b$, $c$ and $d$, in the fuzzy number $A_{1}$ and $A_{2}$, in table \ref{tab1} and figure \ref{fig2}, is not exactly 508. This is because the implemented algorithm adapts the segment length, such that the lowest value in the dataset always appears as the left bound of the crisp interval, and the highest value in the dataset always appears as the right bound of the crisp interval. From table \ref{tab1} and figure \ref{fig2} it can be seen that the lowest of the four values (i.e. 13055), appears as the lower bound of the first crisp interval, $\mu_{1}$, and the highest value (i.e. 14696), appears as the upper bound in the second crisp interval, $u_{2}$. Normally these values cannot be matched precisely without adjusting the segment length, due to rounding errors occurring as a result of equation \ref{eq1} and \ref{eq4}.

We are now able to fuzzify the first four historical enrollments according to membership functions $A_{1}$ and $A_{2}$, defined by:

\[
A_{1} =
\begin{cases}
0, & x < 12547 \\
\dfrac{x - 12547}{13055 - 12547}, & 12547 \leq x \leq 13055 \\
1, & 13055 \leq x \leq 13602 \\
\dfrac{14149 - x}{14149 - 13602}, & 13602 \leq x \leq 14149 \\
0, & x > 14149.
\end{cases}
\]
\begin{center} 
and 
\end{center}
\[
A_{2} =
\begin{cases}
0, & x < 13602 \\
\dfrac{x - 13602}{14149 - 13602}, & 13602 \leq x \leq 14149 \\
1, & 14149 \leq x \leq 14696 \\
\dfrac{15204 - x}{15204 - 14696}, & 14696 \leq x \leq 15204 \\
0, & x > 15204.
\end{cases}
\]

\newpage

\begin{figure}[ht]
	\centering
		\includegraphics[scale=0.7]{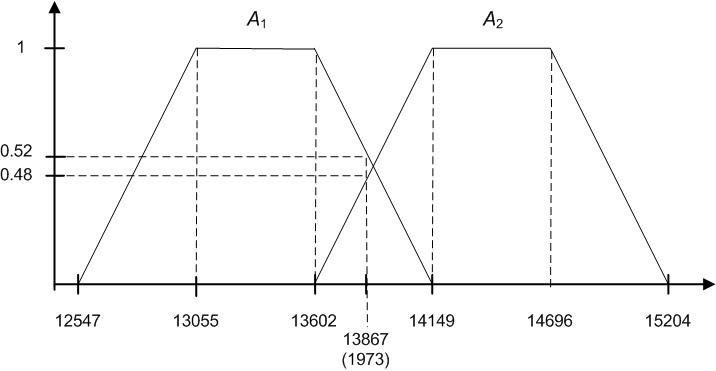}
	\caption{Generated membership functions.}
	\label{fig3}
\end{figure}

As figure \ref{fig3} indicates, the boundaries of $A_{1}$ and $A_{2}$ overlap such that more than one interval may be met. For example, the enrollment for the year 1973 is 13867. This value meets both membership functions. The membership degree in $A_{1}$ is $0.5155 \approx 0.52$, and in $A_{2}$, it is $0.4845 \approx 0.48$. Hence the enrollment for 1973 is fuzzified as $A_{1}$. A special case occurs if the membership degree is 0.5, as this implies that a value has the same membership status in two different sets. In such cases, the respective value is associated with both $A_{1}$ and $A_{2}$. The results of fuzzifying the first four years of enrollment are shown in table \ref{tab2}.

\begin{table}[ht]
	\centering
	\footnotesize
		\begin{tabular}{cccc}		
			\hline
			\parbox[c][0.75cm][c]{2cm}{\centering \textbf {Year}} & \parbox[c][0.75cm][c]{2cm}{\centering \textbf{Enrollment}} & \parbox[c][0.75cm][c]{2cm}{\centering \textbf{Fuzzy set}} & 	\parbox[c][1cm][c]{2.5cm}{\centering \textbf{Membership degree}} \\
			\hline
			1971 & 13055 & $A_{1}$ & 1\\
			1972 & 13563 & $A_{1}$ & 1\\
			1973 & 13867 & $A_{1}$ & 0.52\\
			1974 & 14696 & $A_{2}$ & 1 \\
			\hline
		\end{tabular}
	\caption{Fuzzified Enrollments 1971 - 1974.}
	\label{tab2}
\end{table}

\newpage

\subsubsection{Fuzzifying the Enrollment Dataset}
\label{subsec:fuzzifyingtheenrollmentdataset}
Following the same procedure as in the example from the previous section, the resultant trapezoidal sets, derived by processing the entire enrollment data from 1971 to 1992, are as shown in table \ref{tab3}. The fuzzified annual enrollments are listed in table \ref{tab4}. 

Generally it is assumed that the fuzzy sets, $A_{1}, A_{2},\ldots, A_{n}$, individually represent some linguistic variable. However, with 17 intervals, linguistic values may not make much sense. In the procedure proposed here, this issue is ignored, since the utility of linguistic values has yet to be demonstrated in a fuzzy time series context. However it does not mean that they are not useful in other application contexts.

\begin{table}[ht]
	\centering
	\footnotesize
		\begin{tabular}{|c|c|}
			\hline
			\parbox[c][0.75cm][c]{2.5cm}{\centering \textbf {Fuzzy Set}} & 
			\parbox[c][0.75cm][c]{5cm}{\centering \textbf{Fuzzy Number}}\\
			\hline
			$A_{1}$ & (12861,13055,13245,13436)\\
			\hline
			$A_{2}$ & (13245,13436,13626,13816)\\
			\hline
			$A_{3}$ & (13626,13816,14007,14197)\\
			\hline
			$A_{4}$ & (14007,14197,14388,14578)\\
			\hline
			$A_{5}$ & (14388,14578,14768,14959)\\
			\hline
			$A_{6}$ & (14768,14959,15149,15339)\\
			\hline
			$A_{7}$ & (15149,15339,15530,15720)\\
			\hline
			$A_{8}$ & (15530,15720,15910,16101)\\
			\hline
			$A_{9}$ & (15910,16101,16291,16482)\\
			\hline
			$A_{10}$ & (16291,16482,16672,16862)\\
			\hline
			$A_{11}$ & (16672,16862,17053,17243)\\
			\hline
			$A_{12}$ & (17053,17243,17433,17624)\\
			\hline
			$A_{13}$ & (17433, 17624,17814,18004)\\
			\hline
			$A_{14}$ & (17814, 18004,18195,18385)\\
			\hline
			$A_{15}$ & (18195,18385,18576,18766)\\
			\hline
			$A_{16}$ & (18576,18766,18956,19147)\\
			\hline
			$A_{17}$ & (18956,19147,19337,19531)\\
			\hline
		\end{tabular}
	\caption{Generated fuzzy sets.}
	\label{tab3}
\end{table}
\newpage

 \begin{table}[ht]
	\centering
	\footnotesize
		\begin{tabular}{|c|c|c|}
			\hline
			\parbox[c][0.75cm][c]{2.5cm}{\centering \textbf {Year}} & \parbox[c][0.75cm][c]{2.5cm}{\centering \textbf{Enrollment}} & \parbox[c][0.75cm][c]{2.5cm} {\centering \textbf{Fuzzy Set}} \\[0.5ex]
			\hline
			1971 & 13055 & $A_{1}$\\
			\hline
			1972 & 13563 & $A_{2}$ \\
			\hline
			1973 & 13867 & $A_{3}$ \\
			\hline
			1974 & 14696 & $A_{5}$ \\
			\hline
			1975 & 15460 & $A_{7}$ \\
			\hline
			1976 & 15311 & $A_{7}$\\
			\hline
			1977 & 15603 & $A_{7}$\\
			\hline
			1978 & 15861 & $A_{8}$ \\
			\hline
			1979 & 16807 & $A_{11}$ \\
			\hline
			1980 & 16919 & $A_{11}$\\
			\hline
			1981 & 16388 & $A_{10}$\\
			\hline
			1982 & 15433 & $A_{7}$\\
			\hline
			1983 & 15497 & $A_{7}$\\
			\hline
			1984 & 15145 & $A_{6}$\\
			\hline
			1985 & 15163 & $A_{6}$\\
			\hline
			1986 & 15984 & $A_{8}$\\
			\hline			
			1987 & 16859 & $A_{11}$\\
			\hline
			1988 & 18150 & $A_{14}$\\
			\hline
			1989 & 18970 & $A_{16}$\\
			\hline
			1990 & 19328 & $A_{17}$\\
			\hline
			1991 & 19337 & $A_{17}$\\
			\hline
			1992 & 18876 & $A_{16}$\\
			\hline
		\end{tabular}
	\caption{Fuzzified annual enrollments.}
	\label{tab4}
\end{table}

\subsection{Defuzzifying Output}
In the following section, we will present a novel approach to defuzzify forecast output. The training phase comprises the following steps: 
\newline

\begin{enumerate}
	\item Establish fuzzy set groups (FSG's).
	\item Convert the FSG's into corresponding if rules.
	\item Complete training of the if rules.
	\item Derive forecasts.
\end{enumerate}

Before we go into detail with the individual steps, it is important to understand how
defuzzified output is computed. First, recall that the definition of an $n$-order
fuzzy relationship \cite{Chen2} is denoted as 

\[
F(t-n),\ldots,F(t-2),F(t-1) \rightarrow F(t), 
\]

\noindent where $F$ represents a fuzzified forecast value at time $t$. Normally it is assumed that the left-hand side of the fuzzy
relation is fuzzified in the same manner as the right-hand side. For example, if $F$, on the left-hand side represents a trapezoidal set, then $F$, on the right-hand, side represents a trapezoidal set as well. In the modified version introduced here, this notion has been revised such that $F(t)$ is given by the following defuzzification operator, $Y(t)$, defined by

\begin{equation}
\label{eq7}
Y(t) = \sum_{i = 1}^{n} a_{t-i} \cdot w_{i},
\end{equation}

\noindent where $w_{i} \in [0,1]$ and $a_{t-i}$ denotes the actual value at time $t - i$. Otherwise stated, the defuzzified output is the weighted
sum of the actual values from time $t-n$ to $t-1$, where $n$ depends on the time series span. For
example, if $n$ = 2, we have

\begin{equation}
\label{eq8}
Y(t) = (a_{t-1} \cdot w_{1}) + (a_{t-2} \cdot w_{2}).
\end{equation}

One question that needs to be addressed is how the defuzzification operator deployed here should
be interpreted from a fuzzy logical point of view. The thought here is simply to consider the weights
as a fuzzy relationship between past values (inputs) and the future value (output). Each $w_{i}$ represents
the strength of the causal relationship between a given input and some unknown output. The closer
$w_{i}$ is to 1, the stronger the relationship and vice versa.

It has to be stressed that the defuzzification operator introduced here is not an aggregation
operator \cite{Detyniecki1} from a traditional point of view, since it does not satisfy the boundary condition which is one of the basic conditions of fuzzy aggregation operators. The proposed operator has been specifically adapted to solve the problem at hand because no other operators have been found useful in this context. Averaging operators \cite{Detyniecki1}, for example, never produce outputs less than the minimum value of arguments or larger than the maximum value of arguments. In the current situation, this requirement is undesirable due to the fact that future demand patterns often fluctuate beyond the boundaries of previous min and max values. To illustrate this, we need to take a closer look at the enrollment data
in table \ref{tab4}. For $t = 1973$ and $n = 2$, we get, $a_{1972} = 13563$ and $a_{1971} = 13055$. Assuming $Y(t)$ is an averaging operator, the output is restricted to the interval [13055,13563]. However actual output for $t = 1973$ is 13867 which is out of reach by any averaging operator. Consider another case for $t = 1981$ and $n = 2$. We then get $a_{1980} = 16919$ and $a_{1979} = 16807$. If $Y(t)$ is the min operator, we get min($a_{1980}, a_{1979}$) = 16807, and, if $Y(t)$ is the max operator, we get max($a_{1980}, a_{1979}$) = 16919. But the actual output for $t = 1981$ is 16388 which also is unreachable by any averaging operator. As a consequence, a basic requirement for the defuzzification operator proposed here is that it covers a broader interval than min and max. A reasonable assumption with regards to the bounds of arguments, $a_{t - i}$, is that they are within the limits of the universe set.

\subsection{Establishment of Fuzzy Set Groups}
\label{sec3}

In conventional fuzzy time series, fuzzy relationships are identified immediately after data have been
fuzzified. However, in the model presented here, the right-hand side of the fuzzy relation is not
known until the weights have been determined. Instead of identifying relationships and
establishing fuzzy logical relationship groups, we establish fuzzy set groups (FSG's). The purpose of the FSG establishment
is to partition historical data into unique sets of sub-patterns which subsequently are converted into
IF rules. In the first pass of the algorithm, consecutive sets are grouped
pairwise. Table \ref{tab5} shows the fuzzified data in table \ref{tab4} grouped in this manner. Every FSG appears in
chronological order.

\begin{table}[ht]
	\centering
	\footnotesize	
		\begin{tabular}{|c|c|c|c|}
			\hline
			\parbox[c][0.75cm][c]{2.3cm} { \centering \textbf{Label}}  & \parbox[c][0.75cm][c]{2.3cm} {\centering \textbf{FSG}} & \parbox[c][0.75cm][c]{2.3cm} { \centering \textbf{Label}} & \parbox[c][0.75cm][c]{2.3cm}{\centering \textbf{FSG}} \\
\hline
			1 & $\left\{A_{1},A_{2}\right\}$ & 12  & $\left\{A_{7},A_{7}\right\}$\\
			\hline
			2 & $\left\{A_{2},A_{3}\right\}$ & 13 & $\left\{A_{7},A_{6}\right\}$\\
			\hline
			3 & $\left\{A_{3},A_{5}\right\}$ & 14 & $\left\{A_{6},A_{6}\right\}$\\
			\hline
			4 & $\left\{A_{5},A_{7}\right\}$ & 15 & $\left\{A_{6},A_{8}\right\}$\\
			\hline												
			5 & $\left\{A_{7},A_{7}\right\}$ & 16 & $\left\{A_{8},A_{11}\right\}$\\
			\hline
			6 & $\left\{A_{7},A_{7}\right\}$ & 17 & $\left\{A_{11},A_{14}\right\}$\\
			\hline			
			7 & $\left\{A_{7},A_{8}\right\}$ & 18 & $\left\{A_{14},A_{16}\right\}$\\
			\hline			
			8 & $\left\{A_{8},A_{11}\right\}$ & 19 & $\left\{A_{16},A_{17}\right\}$\\
			\hline			
			9 & $\left\{A_{11},A_{11}\right\}$ & 20 & $\left\{A_{17},A_{17}\right\}$\\
			\hline			
			10 & $\left\{A_{11},A_{10}\right\}$ & 21 & $\left\{A_{17},A_{16}\right\}$\\
			\hline			
			11 & $\left\{A_{10},A_{7}\right\}$ & Ø & Ø \\
			\hline						
		\end{tabular}
	\caption{Establishment of FSG's.}
	\label{tab5}
\end{table}

To exemplify the principles of grouping, consider the years 1971, 1972, and 1973 which are fuzzified as $A_{1}, A_{2}$ and $A_{3}$, respectively. The pairwise grouping of sets is carried out in the
following order:
\[
\left\{F(t-2),F(t-1)\right\}= \left\{A_{i,t-2}, A_{i,t-1}\right\}.	
\]

Following this principle, the following two FSG's are derived:

\[
\left\{F(1971), F(1972)\right\}= \left\{A_{1}, A_{2} \right\} 
\]
\begin{center}
and
\end{center}
\[
\left\{F(1972), F(1973) \right\}=\left\{A_{2}, A_{3}\right\}.
\]

In table \ref{tab5}, these groups are labeled as 1 and 2, respectively. Ultimately the goal of grouping sets in this manner is to obtain a series of FSG's free of ambiguities. An ambiguity occurs if two or more FSG's contain the same combination of fuzzy sets. From table \ref{tab5}, it can be seen that not all FSG's are unique. Note that the FSG's labeled as 5, 6 and 12 are identical, as is the case with 8 and 16. In order to obtain a series of disambiguated FSG's, we extend the ambiguous FSG's by including the previous set in the respective time series. Formally this means that the combination ,$\left\{F(t-2), F(t-1)\right\}$, is extended to include $F(t-3)$, so the respective FSG now equals $\left\{F(t-3), F(t-2), F(t-1)\right\}$. Table \ref{tab6} shows the extensions of the ambiguous FSG's identified in table \ref{tab5}.

\begin{table}[htbp]
	\centering
	\footnotesize
		\begin{tabular}{|c|c|c|c|}
			\hline
			\parbox[c][1.2cm][c]{2cm} {\centering \textbf{Label}}  
			& \parbox[c][1.2cm][c]{3.4cm} {\centering \textbf{\textit{F}(\textit{t}-2), \textit{F}(\textit{t}-1)}} 
			& \parbox[c][1.2cm][c]{2.2cm} {\centering \textbf{\textit{F}(\textit{t}-3)}}
			& \parbox[c][1.2cm][c]{4.6cm}{\centering \textbf{\textit{F}(\textit{t}-3), \textit{F}(\textit{t}-2), \textit{F}(\textit{t}-3)}} \\
\hline
			5 & $\left\{A_{7},A_{7}\right\}$ & $A_{5}$  & $\left\{A_{5},A_{7},A_{7}\right\}$\\
			\hline
			6 & $\left\{A_{7},A_{7}\right\}$ & $A_{7}$ & $\left\{A_{7},A_{7},A_{7}\right\}$\\
			\hline
			8 & $\left\{A_{8},A_{11}\right\}$ & $A_{7}$ & $\left\{A_{7},A_{8},A_{11}\right\}$\\
			\hline
			12 & $\left\{A_{7},A_{7}\right\}$ & $A_{10}$ & $\left\{A_{10},A_{7},A_{7}\right\}$\\
			\hline
			16 & $\left\{A_{8},A_{11}\right\}$ & $A_{6}$ & $\left\{A_{6},A_{8},A_{11}\right\}$\\
			\hline															
		\end{tabular}
	\caption{Extending ambiguous FSG's.}
	\label{tab6}
\end{table}

The extension process is continued until a unique combination of sets is obtained for each
FSG. From table \ref{tab6}, we see that only a single extension is required to obtain a unique combination
of sets in this particular case. An updated overview of the FSG's is shown in table \ref{tab7}.


\begin{table}[htbp]
	\centering
	\footnotesize
		\begin{tabular}{|c|c|c|c|}
			\hline
			\parbox[c][0.75cm][c]{2.3cm} { \centering \textbf{Label}}  & \parbox[c][0.75cm][c]{2.3cm} { \centering \textbf{FSG}} & \parbox[c][0.75cm][c]{2.3cm} { \centering \textbf{Label}} & \parbox[c][0.75cm][c]{2.3cm}{ \centering \textbf{FSG}} \\
\hline
			1 & $\left\{A_{1},A_{2}\right\}$ & 12  & $\left\{A_{10},A_{7},A_{7}\right\}$\\
			\hline
			2 & $\left\{A_{2},A_{3}\right\}$ & 13 & $\left\{A_{7},A_{6}\right\}$\\
			\hline
			3 & $\left\{A_{3},A_{5}\right\}$ & 14 & $\left\{A_{6},A_{6}\right\}$\\
			\hline
			4 & $\left\{A_{5},A_{7}\right\}$ & 15 & $\left\{A_{6},A_{8}\right\}$\\
			\hline												
			5 & $\left\{A_{5},A_{7},A_{7}\right\}$ & 16 & $\left\{A_{6},A_{8},A_{11}\right\}$\\
			\hline
			6 & $\left\{A_{7},A_{7},A_{7}\right\}$ & 17 & $\left\{A_{11},A_{14}\right\}$\\
			\hline			
			7 & $\left\{A_{7},A_{8}\right\}$ & 18 & $\left\{A_{14},A_{16}\right\}$\\
			\hline			
			8 & $\left\{A_{7},A_{8},A_{11}\right\}$ & 19 & $\left\{A_{16},A_{17}\right\}$\\
			\hline			
			9 & $\left\{A_{11},A_{11}\right\}$ & 20 & $\left\{A_{17},A_{17}\right\}$\\
			\hline			
			10 & $\left\{A_{11},A_{10}\right\}$ & 21 & $\left\{A_{17},A_{16}\right\}$\\
			\hline			
			11 & $\left\{A_{10},A_{7}\right\}$ & Ø & Ø \\
			\hline						
		\end{tabular}
	\caption{Disambiguated FSG's.}
	\label{tab7}
\end{table}

\subsection{Converting FSG's into Forecast Rules}
\label{sec4}

Defuzzified output, $Y(t)$, is obtained by matching historical patterns with a corresponding if-then
rule which is generated on the basis of the content of the FSG's. The task is fairly simple as the sequence of elements in each FSG is the same as they appear in time. This means that for any FSG of size $n$, the elements appear in the same sequence as in the corresponding time series:
$F(t-n), F(t-n+1), \ldots,F(t-1)$. Each FSG can be therefore easily transformed into if rules of the form:
\newline \newline \indent
$\textrm{if} (F(t-1)=A_{i,t-1} \wedge F(t-2) = A_{i,t-2} \wedge\ldots\wedge F(t-n) = A_{i,t-n});$
\newline \indent
$\textrm{then } w_{1,t-1} = ?$, $w_{2,t-2} = ?$,\ldots,$w_{n,t-n} = ?$
\newline

For convenience, the sequence of conditions in the if rules appear in reversed order
compared to their equivalent FSG's. For example, an FSG of the form:
\begin{center}
$\left\{A_{i,t-2}, A_{i,t-1}\right\}$,
\end{center}
\begin{center}
is converted into an equivalent if rule of the form:
\end{center}
\begin{center}
$\textrm{if}(F(t-1) = A_{i,t-1} \wedge F(t-2) = A_{i, t-2})$.
\end{center}

Whenever a rule is matched, the resultant weights are returned and the forecast value, $Y(t)$, is
computed according to equation \ref{eq7}. To illustrate this, suppose we need to find a matching if-then rule
when forecasting the enrollment for the year 1973. From table \ref{tab4}, we get $F(1971) = A_{1}$ and
$F(1972) = A_{2}$ for $t = 1973$. Now, assume the following if-then rule exists in the current rule
base:
\begin{center}
$\textrm{if}(F(t-1) = A_{i,t-1} \wedge F(t-2) = A_{i,t-2}) \rightarrow w_{1,t-1} = 0.6488$, $w_{2,t-2}=0.3882.$
\end{center}

The respective if rule is then matched as
\begin{center}
$\textrm{if} (F(1973-1) = A_{1} \wedge F(1973-2) = A_{1}) \rightarrow w_{1,1972} = 0.6488, w_{2,1971} = 0.3882.$
\end{center}

Using equation \ref{eq7}, forecast enrollment for the year 1973 is computed as
\[
Y(1973) = (13563 \cdot 0.6488)+(13055 \cdot 0.3882)= 13867.62 \approx 13868.
\]

By processing all of the FSG's in table \ref{tab7}, a series of incomplete if statements are generated as
shown in table \ref{tab8}. 

\begin{table}[htbp]
	\centering
	\footnotesize
		\begin{tabular}{|c|l|}
			\hline
			\parbox[c][0.75cm][c]{2.3cm} { \centering \textbf{Rule}}  & \parbox[c][0.75cm][c]{9cm} {\centering \textbf{Matching Part}} \\
			\hline
			1 & if$(F(t-1) = A_{2}) \wedge F(t-2) = A_{1})$ \\
			\hline
			2 & if$(F(t-1) = A_{3} \wedge F(t-2) = A_{2})$ \\
			\hline
			3 & if$(F(t-1) = A_{5} \wedge F(t-2) = A_{3})$ \\
			\hline
			4 & if$(F(t-1) = A_{7} \wedge F(t-2) = A_{5})$ \\
			\hline												
			5 & if$(F(t-1) = A_{7} \wedge F(t-2) = A_{7} \wedge F(t-3) = A_{5})$ \\
			\hline
			6 & if$(F(t-1) = A_{7} \wedge F(t-2) = A_{7} \wedge F(t-3) = A_{7})$ \\
			\hline			
			7 & if$(F(t-1) = A_{8} \wedge F(t-2) = A_{7})$ \\
			\hline			
			8 & if$(F(t-1) = A_{11} \wedge F(t-2) = A_{8} \wedge F(t-3) = A_{7})$ \\
			\hline			
			9 & if$(F(t-1) = A_{11} \wedge F(t-2) = A_{11})$ \\
			\hline			
			10 & if$(F(t-1) = A_{10} \wedge F(t-2) = A_{11})$ \\
			\hline			
			11 & if$(F(t-1) = A_{7} \wedge F(t-2) = A_{10})$\\
			\hline
			12 & if$(F(t-1) = A_{7} \wedge F(t-2) = A_{7} \wedge F(t-3) = A_{10})$ \\
			\hline
			13 & if$(F(t-1) = A_{6} \wedge F(t-2) = A_{7})$ \\
			\hline
			14 & if$(F(t-1) = A_{6} \wedge F(t-2) = A_{6})$ \\
			\hline 
			15 & if$(F(t-1) = A_{8} \wedge F(t-2) = A_{6})$ \\
			\hline
			16 & if$(F(t-1) = A_{11} \wedge F(t-2) = A_{8} \wedge F(t-3) = A_{6})$ \\
			\hline
			17 & if$(F(t-1) = A_{14} \wedge F(t-2) = A_{11})$ \\
			\hline
			18 & if$(F(t-1) = A_{16} \wedge F(t-2) = A_{14})$ \\
			\hline
			19 & if$(F(t-1) = A_{17} \wedge F(t-2) = A_{16})$ \\
			\hline
			20 & if$(F(t-1) = A_{17} \wedge F(t-2) = A_{17})$ \\
			\hline
			21 &	if$(F(t-1) = A_{16} \wedge F(t-2) = A_{17})$ \\
			\hline					
		\end{tabular}
	\caption{Generated if rules in chronological order.}
	\label{tab8}
\end{table}

\newpage

\subsection{Training the If-Then Rules With PSO}
\label{sec5}
In the following, we are going to provide an example of how PSO is utilized to tune the weights
in the defuzzification operator in equation \ref{eq7}. The user-defined parameters are set as follows:

\begin{itemize}
	\item The inertial coefficient, $\omega$, equals 1.4.
	\item The self confidence and social confidence coefficient, $c_1$ and $c_2$ both equals 2, respectively.
	\item The minimum and maximum velocity is limited to [-0.01,0.01].
	\item The minimum and maximum position is limited to [0,1].
	\item The number of particles equals 5.
\end{itemize}

The fitness function employed here is the squared error ($SE$), defined by

\begin{equation}
\label{eq11}
SE=(forecast\_value_t - actual\_value_t)^2	
\end{equation}

Basically, the idea is to evaluate the aggregated result, $Y(t)$, against the actual outcome at time
$t$, and adjust the weights in the defuzzification operator such that $SE$ is minimized. By
minimizing $SE$ for each $t$, $MSE$ is minimized as well. In the following example, the stopping
criteria is defined by setting the minimum $SE$ to 3 and the maximum number of iterations to 500.

During the first step of the algorithm, the weights (positions) are initialized. We assume
the existence of a stronger relationship between actual output and the more recent observations in
the time series data. So, if $F(t-1)$ is fuzzified as $A_i$ and $F(t-2)$ as $A_j$, a stronger relationship is
assumed to exist between $A_i$ and $Y(t)$ than between $A_j$ and $Y(t)$. Therefore, relatively higher
weights are assigned to the most recent observations when positions are initialized. Applying this
approach, $w_{t-i}$ will usually remain larger than $w_{t-i+1}$ at the point of termination. Table \ref{tab9} and \ref{tab10}
respectively show the initial positions and velocities of all particles for a given rule.


\begin{table}[ht]
	\centering
	\footnotesize
		\begin{tabular}{|c|c|c|c|}
			\hline
			\parbox[c][1.1cm][t]{2.5cm} {\centering \textbf{\\ Particle}}  & \parbox[c][1.1cm][t]{2.5cm} {\centering \textbf{\\ Position 1} \\ $\boldsymbol{\left(w_{1}\right)}$} & 	
			\parbox[c][1.1cm][t]{2.5cm} {\centering \textbf{\\ Position 2} \\ $\boldsymbol{\left(w_{2}\right)}$} & \parbox[c][1.1cm][t]{2.5cm} {\centering \textbf{\\ SE}}\\
			\hline
			1 & 0.75 & 0.5 & 8,024,473 \\
			\hline
			2 & 0.75 & 0.5 & 8,024,473 \\
			\hline
			3 & 0.75 & 0.5 & 8,024,473 \\
			\hline
			4 & 0.75 & 0.5 & 8,024,473 \\
			\hline												
			5 & 0.75 & 0.5 & 8,024,473 \\
			\hline
		\end{tabular}
	\caption{Initial positions of all particles.}
	\label{tab9}
\end{table}

In the current example, rule 1 in table \ref{tab8} is trained. As can be seen from table \ref{tab9}, the personal best
positions are the same for all particles at the initialization phase. Hence the personal best position equals
the global best position for all particles.

\begin{table}[ht]
	\centering
	\footnotesize
		\begin{tabular}{|c|c|c|}
			\hline
			\parbox[c][0.75cm][c]{3cm} {\centering \textbf{Particle}}  & \parbox[c][0.75cm][c]{3cm} {\centering $\boldsymbol{v_{1}}$} 
			& \parbox[c][0.75cm][c]{3cm} {\centering $\boldsymbol{v_{2}}$} \\
			\hline
			1 & 0.0049 & 0.0011 \\
			\hline
			2 & 0.0032 & 0.0065 \\
			\hline
			3 & 0.0034 & 0.0081 \\
			\hline
			4 & 0.0023 & 0.0009 \\
			\hline												
			5 & 0.0007 & 0.0048 \\
			\hline			
		\end{tabular}
	\caption{Randomized initial velocities of all particles.}
	\label{tab10}
\end{table}

When all particles and velocities have been initialized, the velocities are updated before positions
are incremented. Velocities are updated according to equation \ref{eq9}. The computations yield:

\begin{table}[ht]
	\centering
	\footnotesize
		\begin{tabular}{ll}
$v_{1,1} = (1.4 \cdot 0.0049) + 2 \cdot r_1(0.75-0.75)+2 \cdot r_2(0.75-0.75)$ & = 0.0069\\
$v_{1,2} = (1.4 \cdot 0.0011) + 2 \cdot r_1(0.5-0.5)+2 \cdot r_2(0.5-0.5)$ & = 0.0015\\			
$v_{2,1} = (1.4 \cdot 0.0032) + 2 \cdot r_1(0.75-0.75)+2 \cdot r_1(0.75-0.75)$ & = 0.0045\\			
$v_{2,2} = (1.4 \cdot 0.0065) + 2 \cdot r_1(0.5-0.5)+2 \cdot r_2(0.5-0.5)$ & = 0.0091\\			
$v_{3,1} = (1.4 \cdot 0.0034) + 2 \cdot r_1(0.75-0.75)+2 \cdot r_2(0.75-0.75)$ & = 0.0048\\
$v_{3,2} = (1.4 \cdot 0.0081) + 2 \cdot r_1(0.5-0.5)+2 \cdot r_2(0.5-0.5)$ & = 0.0113\\			
$v_{4,1} = (1.4 \cdot 0.0023) + 2 \cdot r_1(0.75-0.75)+2 \cdot r_2(0.75-0.75)$ & = 0.0032\\			
$v_{4,2} = (1.4 \cdot 0.0009) + 2 \cdot r_1(0.5-0.5)+2 \cdot r_2(0.5-0.5)$ & = 0.0013\\		
$v_{5,1} = (1.4 \cdot 0.0007) + 2 \cdot r_1(0.75-0.75)+2 \cdot r_2(0.75-0.75)$ & = 0.0001\\
$v_{5,2} = (1.4 \cdot 0.0048) + 2 \cdot r_1(0.5-0.5)+2 \cdot r_2(0.5-0.5)$ & = 0.0067.		
	\end{tabular}
\label{tab11}	
\end{table}
\newpage

Positions are incremented according to equation \ref{eq10}. Incremented positions after the first
iteration are shown in table \ref{tab12}.

\begin{table}[ht]
	\centering
	\footnotesize
		\begin{tabular}{|c|c|c|c|}
			\hline
			\parbox[c][1cm][c]{2cm} {\centering \textbf{Particle}}  & \parbox[c][1cm][c]{2.5cm} {\centering $\boldsymbol{w_{1}}$} & 	
			\parbox[c][1cm][c]{2.5cm} {\centering $\boldsymbol{w_{2}}$} & \parbox[c][1cm][c]{2cm} {\centering \textbf{SE}}\\
			\hline
			1 & 0.7549 & 0.5011 & 8,488,885 \\
			\hline
			2 & 0.7532 & 0.5065 & 8,767,574 \\
			\hline
			3 & 0.7534 & 0.5081 & 8,907,895 \\
			\hline
			4 & 0.7523 & 0.5009 & 8,269,618 \\
			\hline												
			5 & 0.7507 & 0.5048 & 8,438,491 \\
			\hline
		\end{tabular}
	\caption{The positions of all particles after the first iteration.}
	\label{tab12}
\end{table}

After the first iteration, none of the computed $SE$ values in table \ref{tab12} are less than 8,024,473. Thus no
personal best positions nor global best positions are reached at this point. At some point, the
stopping criteria are met and the algorithm terminates. The personal best positions of all particles
after termination are listed in table \ref{tab13}.

\begin{table}[ht]
	\centering
	\footnotesize
		\begin{tabular}{|c|c|c|c|}
			\hline
			\parbox[c][1cm][c]{2cm} {\centering \textbf{Particle}}  & \parbox[c][1cm][c]{2.5cm} {\centering $\boldsymbol{w_{1}}$} & 	
			\parbox[c][1cm][c]{2.5cm} {\centering $\boldsymbol{w_{2}}$} & \parbox[c][1cm][c]{2cm} {\centering \textbf{SE}}\\
			\hline
			1 &  0.6738 &  0.3699 &  10,159\\
			\hline
			2 &  0.6854 &  0.3502 &  1\\
			\hline
			3 &  0.6686 &  0.3662 &  325\\
			\hline
			4 &  0.6724 &  0.3482 &  40,597\\
			\hline												
			5 &  0.6879 &  0.3383 &  14,522\\
			\hline
		\end{tabular}
	\caption{Particle positions after termination.}
	\label{tab13}
\end{table}

\newpage

According to table \ref{tab13}, particle 2 has the global best position. Hence the weights associated with the rule in question
equals 0.6854 and 0.3502. The results of training the rules in table \ref{tab8} are shown in table \ref{tab14}.



\begin{algorithm}
\caption{ PSO algorithm for training of the if-then rules.}
\label{algorithm1}
\begin{algorithmic}
\For{all rules}
\State find the matching pattern in the fuzzified dataset
\State retrieves actual values within the range of pattern and initializes variables
\While{stopping criteria is unsatisfied}
\For{each particle $p_{i}$, from $i=1$ to $n$}
\State $v_{ij} = \omega \cdot v_{ij} + c_{1} \cdot r_{1}(local\_b_{ij} - w_{ij}) + c_{2} \cdot r_{2}(global\_b_{j}-w_{ij})$, from $j=1$ to $n$ 
\If{$V_{min} > v_{ij}$} 
\State $v_{ij} = V_{min}$
\EndIf
\If{$V_{max} < v_{ij}$}
\State $v_{ij} = V_{max}$
\EndIf
\State update position as $w_{ij} = w_{ij} + v_{ij}$, from $i = 1$ to $n$
\State compute defuzzified output as $Y(t)_{i} = \sum_{i = 1}^{n} a_{t-j} \cdot w_{ij}$
\State compute squared error $SE_{i} = (Y(t)_{i}-actual\_value_{t})^2$
\If{$SE_{i} < SE_{local\_best}$}
\State $SE_{local\_best} = SE_{i}$
\State $local\_b_{ij} = w_{ij}$, from $i = 1$ to $n$
\EndIf
\If{$SE_{i} < SE_{global\_best}$}
\State $SE_{global\_best} = SE_{i}$
\State $global\_best_{ij} = w_{ij}$, from $i = 1$ to $n$
\EndIf
\EndFor
\EndWhile
\State update weights as \{$w_{t-1} = global\_best_{1} \wedge \ldots \wedge w_{t-n}=global\_best_{n}$\} 
\EndFor
\end{algorithmic}
\end{algorithm}

\begin{sidewaystable} [tbp] 	
		\footnotesize
		\begin{tabular}{|c|l|l|}		
			\hline			
			\parbox[c][0.75cm][c]{1.5cm} {\centering \textbf{Rule}} & 
			\parbox[c][0.75cm][c]{7.7cm} {\centering \textbf{Matching Part}} & 
			\parbox[c][0.75cm][c]{7.7cm} {\centering \textbf{Weights}} \\
			\hline			 
			1 & if$(F(t-1) = A_{2}) \wedge F(t-2) = A_{1})$ & then $w_1$=0.6488 and $w_2$=0.3882\\ 
			\hline			
			2 & if$(F(t-1) = A_{3} \wedge F(t-2) = A_{2})$ & then $w_1$=0.6586 and $w_2$=0.4102\\
			\hline			
			3 & if$(F(t-1) = A_{5} \wedge F(t-2) = A_{3})$ & then $w_1$=0.667 and $w_2$=0.408 \\
			\hline			
			4 & if$(F(t-1) = A_{7} \wedge F(t-2) = A_{5})$ & then $w_1$=0.6395 and $w_2$=0.369 \\
			\hline															
			5 & if$(F(t-1) = A_{7} \wedge F(t-2) = A_{7} \wedge F(t-3) = A_{5})$ & then $w_1$=0.4411, $w_2$=0.3158 and $w_3$ = 0.2699 \\
			\hline			
			6 & if$(F(t-1) = A_{7} \wedge F(t-2) = A_{7} \wedge F(t-3) = A_{7})$ & then $w_1$=0.4638, $w_2$=0.4645 and $w_3$=0.0978 \\
			\hline						
			7 & if$(F(t-1) = A_{8} \wedge F(t-2) = A_{7})$ & then $w_1$=0.6695 and $w_2$=0.3967 \\
			\hline						
			8 & if$(F(t-1) = A_{11} \wedge F(t-2) = A_{8} \wedge F(t-3) = A_{7})$ & then $w_1$=0.4379,$w_2$=0.3892 and $w_3$=0.2171 \\
			\hline						
			9 & if$(F(t-1) = A_{11} \wedge F(t-2) = A_{11})$ & then $w_1$=0.1604 and $w_2$=0.8137 \\
			\hline						
			10 & if$(F(t-1) = A_{10} \wedge F(t-2) = A_{11})$ & then $w_1$=0.5497 and $w_2$=0.3798 \\
			\hline						
			11 & if$(F(t-1) = A_{7} \wedge F(t-2) = A_{10})$ & then $w_1$=0.5997 and $w_2$=0.3809 \\
			\hline			
			12 & if$(F(t-1) = A_{7} \wedge F(t-2) = A_{7} \wedge F(t-3) = A_{10})$ & then $w_1$=0.4151, $w_2$=0.3966 and $w_3$=0.1582 \\
			\hline			
			13 & if$(F(t-1) = A_{6} \wedge F(t-2) = A_{7})$ & then $w_1$=0.6194 and $w_2$=0.3731 \\
			\hline			
			14 & if$(F(t-1) = A_{6} \wedge F(t-2) = A_{6})$ & then $w_1$=0.7524 and $w_2$=0.302 \\
			\hline 			
			15 & if$(F(t-1) = A_{8} \wedge F(t-2) = A_{6})$ & then $w_1$=0.3869 and $w_2$=0.704 \\
			\hline			
			16 & if$(F(t-1) = A_{11} \wedge F(t-2) = A_{8} \wedge F(t-3) = A_{6})$ & then $w_1$=0.4668, $w_2$=0.3847 and $w_3$=0.2725 \\
			\hline			
			17 & if$(F(t-1) = A_{14} \wedge F(t-2) = A_{11})$ & then $w_1$=0.654 and $w_2$=0.4212 \\
			\hline			
			18 & if$(F(t-1) = A_{16} \wedge F(t-2) = A_{14})$ & then $w_1$=0.635 and $w_2$=0.4012 \\
			\hline			
			19 & if$(F(t-1) = A_{17} \wedge F(t-2) = A_{16})$ & then $w_1$=0.6202 and $w_2$=0.3874 \\
			\hline			
			20 & if$(F(t-1) = A_{17} \wedge F(t-2) = A_{17})$ & then $w_1$=0.5932 and $w_2$=0.3831 \\
			\hline
			21 & if$(F(t-1) = A_{16} \wedge F(t-2) = A_{17})$ & then $w_1$=? and $w_2$=? \\
			\hline					
		\end{tabular}		
	\caption{Generated if-then rules in chronological order.}
	\label{tab14}
\end{sidewaystable}
\newpage

\section{Experimental Results}
\label{sec:results}
To make this work comparable with those of others, the procedure for evaluating model performance is carried out in the same manner as in other publications. That is by forecasting historical enrollments and then evaluating performance on the basis of performance indicators. The mean squared error (MSE) and the mean average percentage error (MAPE) are used as the basis for evaluating model performance:

\begin{equation}
\label{eq12}
MSE = \frac{1}{n} \sum_{i=1}^{n} (F_{i}-A_{i})^2
\end{equation}

\begin{equation}
\label{eq13}
MAPE = \frac{1}{n} \sum_{i=1}^{n} \frac {|F_{i}-A_{i}|}{A_{i}} \times 100,
\end{equation}
where $A_{i}$ and $F_{i}$ denote the actual output and forecasted ouput at time $i$, respectively.

A comparison of the proposed model versus other related models is presented in table \ref{tab15}. The model's parameters used in this experiment are the same as in the example in section \ref{sec5}. We have selected the best result out of 10 runs. The models referenced in table \ref{tab15} are all among those with the highest accuracy rates published. As can be seen from table \ref{tab15}, the MSE and the MAPE of the proposed model is 1 and 0.006, respectively. This is significantly lower than for any of the referenced models. Judging by these results, it can therefore be concluded that the proposed model outperforms any fuzzy time series model currently published. 

In the introductory section, it was argued that one of the potential drawbacks of the high-order models is that data becomes underutilized as the model's order increases. To explain this in further detail, we need to take a closer look at the outcome produced by the models in \cite{ChenChung1} and \cite{Kuo1}. Note that the first 7 and 8 years of enrollment data are not forecast. This is because the number of fuzzy sets to be matched for each forecast period increases with the order. For example in \cite{Kuo1}, the number of sets to be matched is 8. By increasing the number of set combinations to be matched, fewer forecast rules are obtained for future use, and thus data becomes underutilized. Moreover, as the number of sets to be matched increases, the statistical likelihood of encountering equivalent pattern combinations in future datasets decreases.

\section{Conclusions}
\label{sec:conclusions}
In this paper, we have introduced a novel fuzzy time series model that integrates the principles of weighted summation and particle swarm optimization (PSO) to individually fine-tune fuzzy rules. This combination of techniques enables more precise calibration of fuzzy rules to align with the underlying data patterns, reducing their sensitivity to chosen interval partitions. Furthermore, the individualized adjustment of fuzzy rules diminishes the necessity for increasing the model's order, in contrast to high-order fuzzy time series models. As a result, we achieve more effective data utilization in two ways: (1) by increasing the number of fuzzy rules and (2) by reducing the number of pattern combinations required to match future time series data. Additionally, we have introduced a fuzzification method capable of determining the appropriate number of interval partitions based on observed variations in the time series data.

Empirical experiments demonstrate that our proposed model outperforms comparable approaches. The practical utility of this method heavily relies on its capacity to derive fitting fuzzy rules. The weighted fuzzy rules exhibit a remarkable ability to closely emulate the original time series. The application of this approach for forecasting future values within the time series is a potential avenue for future research.

\footnotesize
\begin{table}
\begin{center}
\begin{tabular} {|*{10}{c|}}
\hline
	\parbox[c][1cm][t]{1cm} {\centering \textbf {\\ Year}} 
& \parbox[c][1cm][t]{2cm} {\centering \textbf {\\ Enroll \\}} 
& \parbox[c][2cm][t]{1cm} {\centering \textbf {\\Chen \\ (order 3)} \\ \cite{Chen1}} 
& \parbox[c][2cm][t]{1cm} {\centering \textbf {\\Li \\ and \\ Cheng} \\ \cite{LiCheng1}} 
& \parbox[c][2cm][t]{2cm} {\centering \textbf {\\ {Sing} \\ (order 3)} \\ \cite{Sing2}}
& \parbox[c][2cm][t]{2cm} {\centering \textbf {\\ Stevenson \\ and \\ Porter} \\ \cite{Stevenson1}}  
& \parbox[c][2cm][t]{1cm} {\centering \textbf {\\ Chen \\ and \\ Hsu} \\ \cite{ChenHsu1}}
& \parbox[c][2cm][t]{1cm} {\centering \textbf {\\ Chen \\ and \\ Chung \\(order 9)} \\ \cite{ChenChung1} \\} 
& \parbox[c][2cm][t]{1cm} {\centering \textbf {\\ Kuo et al \\ (order 9)} \\ \cite{Kuo1}} 
& \parbox[c][2cm][t]{1.cm} {\centering \textbf {\\ Proposed \\ model}} \\
\hline 1971 & 13055 & - & - & - & - & - & - & - & -\\
\hline 1972 & 13563 & - &  13500 & - &  13410 &  13750 & - & - & - \\
\hline 1973 &  13867 & - &  13500 & - &  13932 &  13875 & - & - &  13868 \\
\hline 1974 &  14696 &  14500 &  14500 &  14750 &  14664 &  14750 & - & - &  14696 \\
\hline 1975 &  15460 &  15500 &  15500 &  15750 &  15423 &  15375 & - & - &  15460 \\
\hline 1976 &  15311 &  15500 &  15500 &  15500 &  15847 &  15313 & - & - &  15309 \\
\hline 1977 &  15603 &  15500 &  15500 &  15500 &  15580 &  15625 & - & - &  15602 \\
\hline 1978 &  15861 &  15500 &  15500 &  15500 &  15877 &  15813 & - & - &  15861 \\
\hline 1979 &  16807 &  16500 &  16500 &  16500 &  16773 &  16834 &  16846 & - &  16806 \\
\hline 1980 &  16919 &  16500 &  16500 &  16500 &  16897 &  16834 &  16846 &  16890 &  16919 \\
\hline 1981 &  16388 &  16500 &  16500 &  16500 &  16341 &  16416 &  16420 &  16395 &  16390 \\
\hline 1982 &  15433 &  15500 &  15500 &  15500 &  15671 &  15375 &  15462 &  15434 &  15434 \\
\hline 1983 &  15497 &  15500 &  15500 &  15500 &  15507 &  15375 &  15462 &  15505 &  15497 \\
\hline 1984 &  15145 &  15500 &  15500 &  15250 &  15200 &  15125 &  15153 &  15153 &  15143 \\
\hline 1985 &  15163 &  15500 &  15500 &  15500 &  15218 &  15125 &  15153 &  15153 &  15163 \\
\hline 1986 &  15984 &  15500 &  15500 &  15500 &  16035 &  15938 &  15977 &  15971 &  15982 \\
\hline 1987 &  16859 &  16500 &  16500 &  16500 &  16903 &  16834 &  16846 &  16890 &  16859 \\
\hline 1988 &  18150 &  18500 &  18500 &  18500 &  17953 &  18250 &  18133 &  18124 &  18150 \\
\hline 1989 &  18970 &  18500 &  18500 &  18500 &  18879 &  18875 &  18910 &  18971 &  18971 \\
\hline 1990 &  19328 &  19500 &  19500 &  19500 &  19303 &  19250 &  19334 &  19337 &  19328 \\
\hline 1991 &  19337 &  19500 &  19500 &  19500 &  19432 &  19250 &  19334 &  19337 &  19336 \\
\hline 1992 &  18876 &  18500 &  18500 &  18750 &  18966 &  18875 &  18910 &  18882 &  18875 \\
\hline
\multicolumn{10}{|l|}{} \\
\hline
\multicolumn{2}{|c|}{ {\centering \textbf{MSE}}} &  86694 &  85040 &  76509 &  21575 &  5611 &  1101 &  234 &  1 \\
\hline
\multicolumn{2}{|c|}{ {\centering \textbf{MAPE}}} &  1.53 &  1.53 &  1.41 &  0.57 &  0.36 &  0.15 &  0.014 &  0.006 \\
\hline
\end{tabular}
\caption{\label{tab15}Comparing results of different fuzzy time series models.}
\end{center}
\end{table}




\bibliographystyle{plain}
\bibliography{ftspso}

\begin{thebibliography}{10}

\bibitem{Chen1}
Shyi-Ming Chen.
\newblock Forecasting enrollments based on fuzzy time series.
\newblock {\em Fuzzy Sets and Systems 81}, 1996.

\bibitem{Chen2}
Shyi-Ming Chen.
\newblock Forecasting enrollments based on high order fuzzy time series.
\newblock {\em Cybernetics and Systems: An Int. Journal 33}, 2002.

\bibitem{ChenChung1}
Shyi-Ming Chen and Nien-Yi Chung.
\newblock Forecasting enrollments using high-order fuzzy time series and genetic algorithms.
\newblock {\em International Journal of Intelligent Systems 21}, 2006.

\bibitem{ChenHsu1}
Shyi-Ming Chen and Chia-Ching Hsu.
\newblock A new method to forecast enrollments using fuzzy time series.
\newblock {\em Int. Journal of Applied Science and Engineering 2}, 2004.

\bibitem{ChenHwang1}
Shyi-Ming Chen and Hwang Jeng-Ren.
\newblock Temperature prediction using fuzzy time series.
\newblock {\em Systems, Management and Cybernetics 30}, 2000.

\bibitem{ChenChengTeoh1}
Tai-Liang Chen, Ching-Hsue Cheng, and Hia-Jong Teoh.
\newblock High-order fuzzy time series based on multi period adaption model for forecasting stock markets.
\newblock {\em Physica A 387}, 2008.

\bibitem{Cheng1}
Ching-Hsue Cheng, Jing-Rong Chang, and Che-An Yeh.
\newblock Entrophy-based and trapezoid fuzzification-based fuzzy time series approaches for forecasting \uppercase{IT} project cost.
\newblock {\em Technological Forecasting \& Social Change 73}, 2006.

\bibitem{Cheng2}
Ching-Hsue Cheng, Guang-Wei Cheng, and Jia-Wen Wang.
\newblock Multi-attribute fuzzy time series method based on fuzzy clustering.
\newblock {\em Expert Systems with Applications 34}, 2008.

\bibitem{Detyniecki1}
Marcin Detyniecki.
\newblock Fundamentals on aggregation operators, 2001.
\newblock \url{http://www-poleia.lip6.fr/~marcin/papers/Detynieck_AGOP_01.pdf}(validated April 2 2011).

\bibitem{HuarngYuHsu1}
Kun-Huang Huarng, Tiffany Hui-Kuang Yu, and Yu~W. Hsu.
\newblock A multivariate heuristic model for fuzzy-time series forecasting.
\newblock {\em Systems, Management and Cybernetics 37}, 2007.

\bibitem{Huarng1}
Kunhuang Huarng.
\newblock Effective lengths on intervals to improve forecasting in fuzzy time series.
\newblock {\em Fuzzy Sets and Systems 123}, 2001.

\bibitem{HuarngYu2}
Kunhuang Huarng and Hui-Kuang Yu.
\newblock A type 2 fuzzy time series model for stock index forecasting.
\newblock {\em Physica A 353}, 2005.

\bibitem{HuarngYu3}
Kunhuang Huarng and Tiffany Hui-Kuang Yu.
\newblock The application of neural networks to forecast fuzzy time series.
\newblock {\em Physica A 363}, 2006.

\bibitem{Huarng2}
Kunhuang Huarng and Tiffany Hui-Kuang Yu.
\newblock Ratio-based lengths of intervals to improve fuzzy time series forecasting.
\newblock {\em Systems Management and Cybernetics 36}, 2006.

\bibitem{ChenHwangLee1}
Jeng-Ren Hwang, Shyi-Ming Chen, and Chia-Hoang Lee.
\newblock Handling forecast problems using fuzzy time series.
\newblock {\em Fuzzy Sets and Systems 100}, 1998.

\bibitem{JilaniBurney1}
Tahseen~A. Jilani and Syed M.~A. Burney.
\newblock A refined fuzzy time series model for stock market forecasting.
\newblock {\em Statistical Mechanics and its Applications 387}, 2008.

\bibitem{JilianiBurneyArdil1}
Tahseen~A. Jiliani and Syed M.~A. Burney.
\newblock Multivariate high order fuzzy time series forecasting for car road accidents.
\newblock {\em Int. Journal of Computational Intelligence 4}, 2008.

\bibitem{KennedyEberhart1}
James Kennedy and Rusell Eberhart.
\newblock Particle swarm optimization.
\newblock {\em Proc. of IEEE Int. Conference on Neural Network}, 1995.

\bibitem{KennedyEberhart2}
James~F. Kennedy, Rusell~C. Eberhart, and Yuhui Shi.
\newblock {\em Swarm intelligence}.
\newblock Morgan Kaufman, 2001.

\bibitem{KlirJuan1}
George~J. Klir and Bo~Yuan.
\newblock {\em Fuzzy sets and fuzzy logic: theory and applications}.
\newblock Prentice Hall, 1995.

\bibitem{Kuo1}
I-Hong Kuo, Shi-Jinn Horng, Tzong-Wann Kao, Tsung-Lieh Lin, Lee Cheng-Ling, and Yi~Pan.
\newblock An improved method for forecasting enrollments based on fuzzy time series and particle swarm optimization.
\newblock {\em Expert Systems with Applications 36}, 2009.

\bibitem{LeeChou1}
Hsun-Shih Lee and Ming-Tao Chou.
\newblock Fuzzy forecasting based on fuzzy time series.
\newblock {\em Int. Journal of Computer Mathematics 81}, 2004.

\bibitem{LiChen1}
Sheng-Tun Li and Yen-Peng Chen.
\newblock Natural partioning-based forecasting model for fuzzy time series.
\newblock {\em Fuzzy Systems 3}, 2004.

\bibitem{LiCheng1}
Sheng-Tun Li and Yi-Chung Cheng.
\newblock Deterministic fuzzy time series model for forecasting enrollments.
\newblock {\em Expert Systems with Applications 34}, 2008.

\bibitem{LiChengLin1}
Sheng-Tun Li, Yi-Chung Cheng, and Lin Su-Yu.
\newblock A fcm-based deterministic forecasting model for fuzzy time series.
\newblock {\em Computers and Mathematics with Applications 56}, 2008.

\bibitem{ortiz2018weighted}
Daniel Ortiz-Arroyo and Jens~Runi Poulsen.
\newblock A weighted fuzzy time series forecasting model.
\newblock {\em Indian journal of science and technology}, 11(27):1--11, 2018.

\bibitem{Sing1}
Shiva~Raj Singh.
\newblock A robust method of forecasting based on fuzzy time series.
\newblock {\em Applied Mathematics and Computation 188}, 2007.

\bibitem{Sing2}
Shiva~Raj Singh.
\newblock A simple time variant method for fuzzy time series forecasting.
\newblock {\em Cybernetics and Systems: An Int. Journal 38}, 2007.

\bibitem{SongChissom1}
Qiang Song and Brad Chissom.
\newblock Forecasting enrollments with fuzzy time series - part i.
\newblock {\em Fuzzy Sets and Systems 54}, 1993.

\bibitem{SongChissom3}
Qiang Song and Brad Chissom.
\newblock Fuzzy time series and its models.
\newblock {\em Fuzzy Sets and Systems 54}, 1993.

\bibitem{SongChissom2}
Qiang Song and Brad Chissom.
\newblock Forecasting enrollments with fuzzy time series - part ii.
\newblock {\em Fuzzy Sets and Systems 62}, 1994.

\bibitem{Stevenson1}
Meredith Stevenson and John~E. Porter.
\newblock Fuzzy time series forecasting using percentage change as the universe of discourse.
\newblock {\em World Academy of Science, Engineering and Technology 55}, 2009.

\bibitem{SullivanWoodall1}
Joe Sullivan and William~H. Woodall.
\newblock A comparison of fuzzy forecasting and markov modeling.
\newblock {\em Fuzzy Sets and Systems 64}, 1994.

\bibitem{TsaiYu1}
Chao-Chih Tsai and Shun-Jyh Wu.
\newblock Forecasting enrollments with high-order fuzzy time series.
\newblock {\em Fuzzy Information Processing Society}, 2000.

\bibitem{Tsaur1}
Ruey-Chyn Tsaur, Jia-Chi Yang, and Hsiao-Fan Wang.
\newblock Fuzzy relation analysis in fuzzy time series model.
\newblock {\em Computers and Mathematics with Applications 49}, 2005.

\bibitem{Yu1}
Hui-Kuang Yu.
\newblock Weighted fuzzy time series models for taiex forecasting.
\newblock {\em Physica A 349}, 2005.

\bibitem{Zadeh1}
Lofti~A. Zadeh.
\newblock Fuzzy sets.
\newblock {\em Information and Control 8}, 1965.

\end{thebibliography}







\end{document}